\documentclass[sigconf]{acmart}
 \usepackage{multirow}
 \usepackage{enumitem}
 \usepackage{subfigure}
 \usepackage{yfonts}
 \usepackage{eufrak}
 \usepackage{bbold}
\usepackage{amsmath, bm}
\usepackage{booktabs}
\usepackage{lipsum}

\usepackage{balance}

\copyrightyear{2020} 
\acmYear{2020} 
\setcopyright{acmcopyright}
\acmConference[MM '20]{Proceedings of the 28th ACM International Conference on Multimedia}{October 12--16, 2020}{Seattle, WA, USA}
\acmBooktitle{Proceedings of the 28th ACM International Conference on Multimedia (MM '20), October 12--16, 2020, Seattle, WA, USA}
\acmPrice{15.00}
\acmDOI{10.1145/3394171.3413661}
\acmISBN{978-1-4503-7988-5/20/10}
   
\begin{document}

\fancyhead{}

\title[Campus3D]{Campus3D: A Photogrammetry Point Cloud Benchmark for Hierarchical Understanding of Outdoor Scene}


\author{Xinke Li$^{1}$, 
        Chongshou Li$^{1,*}$,
        Zekun Tong$^{1}$, 
        Andrew Lim$^{1}$, 
        Junsong Yuan$^{2}$}
\author{Yuwei Wu$^{1}$, 
        Jing Tang$^{1}$, 
        Raymond Huang$^{1}$}
\affiliation{
\institution{$^1$Department of Industrial Systems Engineering and Management, National University of Singapore, Singapore \\ $^2$Department of Computer Science and Engineering, State University of New York at Buffalo, Buffalo, NY, USA
}
}
\email{{xinke.li, zekuntong}@u.nus.edu, {iselc,isealim}@nus.edu.sg, jsyuan@buffalo.edu}
\email{ywwu@u.nus.edu, {isejtang,raymond.huang}@nus.edu.sg}
\thanks{*Corresponding author: Chongshou Li (iselc@nus.edu.sg)}

\begin{abstract}
Learning on 3D scene-based point cloud has received extensive attention as its promising application in many fields, and well-annotated and  multisource  datasets  can catalyze the development of those data-driven approaches.  To facilitate the research of this area,  we present a richly-annotated 3D point cloud dataset for multiple outdoor scene understanding tasks and also an effective learning framework for its hierarchical segmentation task. The dataset was generated via the  photogrammetric processing on unmanned aerial vehicle (UAV) images of the National University of Singapore (NUS) campus,  and has been point-wisely annotated with both hierarchical and instance-based labels.  Based on it,  we formulate a hierarchical learning problem for 3D point cloud segmentation and propose a measurement evaluating consistency across various hierarchies.  To solve this problem, a  two-stage method including multi-task (MT) learning and hierarchical ensemble (HE) with consistency consideration is proposed.  Experimental results  demonstrate the superiority of the proposed method and potential advantages of our hierarchical annotations. In addition, we benchmark results of semantic and instance segmentation, which is accessible online at \url{https://3d.dataset.site} with the dataset and all source codes.
\end{abstract}




\maketitle
\begin{figure}
	\centering		
	\includegraphics[width=1.0\linewidth, height=0.25\textheight]{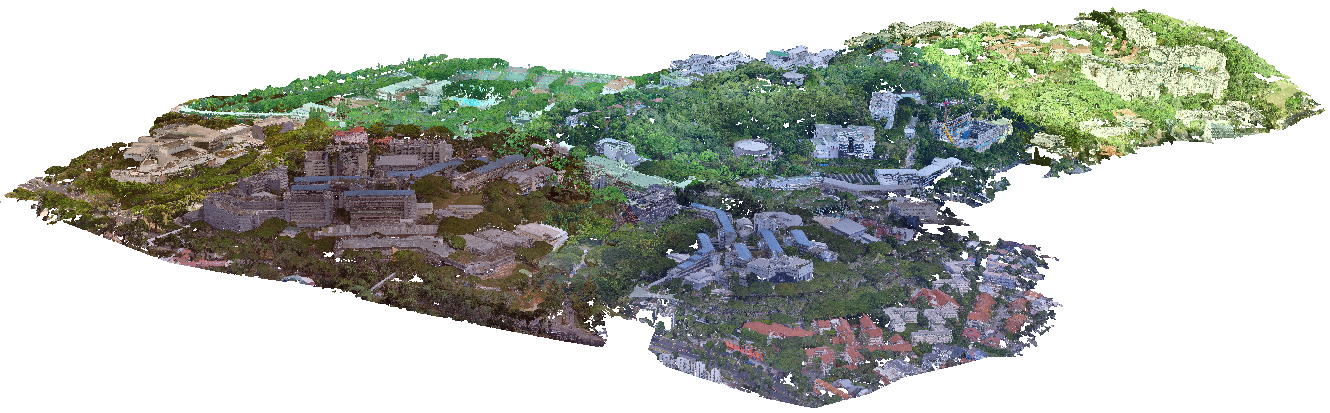} 
	\caption{Overview of six regions of the Campus3D dataset.} \label{Fig:ov_s_r}
	\vspace{-0.6cm}
\end{figure}

\section{Introduction}
Due to the significant progress of 3D sensoring technologies in recent years, multiple sources of 3D point cloud become affordable and easily acquired. Reconstruction of outdoor scene from point cloud has also received an increasing interest, which is critical for various areas such as urban planning and management \cite{carozza2014markerless}, vehicle navigation \cite{cappelle2012virtual}, virtual reality \cite{cirulis20133d} as well as simulation \cite{manyoky2014developing}. As the fundamental step of reconstruction, scene understanding with point cloud data can be greatly facilitated by  
recent advances of machine learning techniques especially the deep learning. Large and well-annotated datasets play a leading role for the successful application of these techniques.

Although dozens of 3D scene-based point cloud datasets are proposed \cite{silberman2012indoor,firman2016rgbd,roynard2018paris,armeni2017joint, serna2014paris,vallet2015terramobilita,hackel2017semantic3d, dai2017scannet, behley2019semantickitti}, majority of them are not perfectly fit for outdoor scene reconstruction.  Firstly, the datasets may face various limitations from their sources which are 
are either RGB-D images  \cite{silberman2012indoor,firman2016rgbd,dai2017scannet, armeni2017joint} or light detection and ranging (LiDAR) based mobile laser scanning (MLS)  \cite{roynard2018paris,serna2014paris,vallet2015terramobilita,behley2019semantickitti} and terrestrial laser scanning (TLS)  \cite{hackel2017semantic3d}.  The RGB-D data can be easily obtained and processed via a mature pipeline \cite{dai2017scannet}, while it is likely prevented from capturing outdoor environment by the limited measurement range. 
The LiDAR scanner usually results in unavoidable severe occlusions and expensive equipment costs although it is good at  
capturing large-scale scenes 
\cite{li2016reconstructing}.  
Secondly, the annotations of extant datasets are not 
targeted  
for outdoor scene reconstruction.  
Following the well-established data format CityGML \cite{kolbe2005citygml}, a standard urban model should contain fine structures of building and other artifacts.  However, such fineness is not presented by current annotations which mainly consist of indoor objects  
or traffic elements   
\cite{behley2019semantickitti}. 
Thus, it is necessary to build new datasets  
with the aim of supporting scene understanding based automatic reconstruction.  

In this work, we construct a photogrametry point cloud dataset Campus3D  from UAV imagery over the National University of Singapore (NUS) campus of 1.58 $\textrm{km}^2$ area. Due to the recent progress of Structure from Motion (SfM),  Multi-View Stereo (MVS) and UAV techniques  \cite{frahm2010building, wu2011multicore},  photogrammetry point cloud is easily accessible from unmanned aerial vehicle (UAV) imagery. This type of data source is able to fulfill the requirement of scene reconstruction because UAV imagery is robust to occlusion, and can effectively obtain the holistic view of the scene.  

Inspired by the multiple levels of details (LoD) in CityGML \cite{kolbe2005citygml} for the reconstruction, we point-wisely annotate this dataset with hierarchical multi-labels for both semantic and instance segmentation.  For a data point, an  example annotations is  construction->building->wall/roof.   The fine-grained label (e.g., wall/roof) can match the LoD2 for  reconstruction \cite{verdie2015lod}, where the building model is detailed to roof and wall structures. For the further study on it,  we organize the labels as a tree with five hierarchical (granularity) levels displayed by Figure \ref{Fig:label_structure}. 
In the end, the whole dataset present a holistic view of scene, which contains 0.94 billion points with 2,530 modality-based instances, 24 semantic classes and 6 pattern-based regions as displayed by Figure \ref{Fig:ov_s_r}.

The proposed dataset with hierarchical annotations is expected to promote better outdoor scene understanding. Based on the constructed label tree, we formulate a hierarchical learning (HL) problem for semantic segmentation, and propose a new metric for consistency across granularity levels named \textit{Consistency Rate} (CR).  Besides accuracy, prediction consistency is an important issue for the HL.  For example,  if  one point is predicted as ``\textit{roof}''  at fine-grained level, the results at the corresponding coarse level must be ``\textit{building}'' and  ``\textit{construction}''(see Figure \ref{Fig:label_structure}), otherwise, it is a violation of the hierarchical relationship. Taking this into consideration, we introduce a two-stage method consist of multi-tasking (MT) learning and hierarchical ensemble (HE). The MT based on neural models jointly learns semantic labeling on different granularity levels. The post-processing HE rigidly ensures the results to fulfill the hierarchical consistency by choosing the most likely root-to-leaf path of the label tree. The results of CR and segmentation task suggest the goodness that the HL method utilizes the hierarchical relationship and the chance that hierarchical annotations assists segmentation tasks.

Furthermore, we establish the benchmarks on the dataset via applying deep models for two classic scene understanding tasks: (1) semantic segmentation and (2) instance segmentation. For the concern of computational efficiency and compatibility to point-based models, we investigate the data prepossess technique and two sampling methods: (1) random block sampling (RBS) and (2) random centered K nearest neighbor (RC-KNN) sampling. And the RBS is chosen as the unified sampling method for benchmarks in view of its better performance.

We summarize the contributions of this paper as follows:
 \begin{itemize} [noitemsep,topsep=0pt]	
	\item A photogrammetry point cloud dataset with hierarchical and instance-based annotations is present. 
	Moreover, an accessible workflow of the acquisition and annotation is provided.
	
	\item An effective two-stage method for the formulated hierarchical semantic segmentation on point cloud is proposed. Experimental results demonstrate the superiority of our HL methods over the non-HL method in terms of both hierarchical consistency and segmentation performance. 
	
	
	\item  We propose new benchmarks for semantic segmentation and instance segmentation on 3D point cloud, and release the source codes\footnote{https://github.com/shinke-li/Campus3D} of the training/evaluation framework as well as the dataset. These benchmarks are standardized with consideration of the unified data prepossess techniques and sampling methods. 
\end{itemize}

\begin{table*}[t]
\small
\vspace{-0.4cm}
	\caption{Comparison between Campus3D and popular scene-based point cloud datasets.}
	\centering	
	\begin{tabular}{l|lllll}
		\toprule
		Dataset & Data Source Type & Area/Length & Scene Type &  Point \#  &  Designed 3D Task   \\
		\midrule
		ScanNet \cite{dai2017scannet} & RGB-D &   floor: 34,453 \scalebox{0.7}{$\textrm{m}^2$}    &   Indoor    &    -    &  Object classification;   \\

		&            &   surface: 78,595 \scalebox{0.7}{$\textrm{m}^2$}    &               &          &     Instance $\&$ semantic segmentation;   \\
		&             &       &             &              &   CAD model retrieval  \\
		\hline
		S3DIS\cite{armeni20163d} & RGB-D &  6000  \scalebox{0.7}{$\textrm{m}^2$} & Indoor & 695.9M & Object detection\\
		\hline
		Matterport3D\cite{chang2017matterport3d} & RGB-D &   floor: 46,561  \scalebox{0.7}{$\textrm{m}^2$}   &    Indoor   &  -     &    Instance $\&$ semantic segmentation       \\
		&            &   surface: 219,398 \scalebox{0.7}{$\textrm{m}^2$}    &               &          &   \\
		\hline
		SemanticKITTI \cite{behley2019semantickitti} & Velodyne HDL-64E (MLS) & 39.2 km & Outdoor & 4,549M &  Semantic segmentation \\
		&             &       &             &              &    Semantic scene completion  \\
		\hline
		Semantic3D\cite{hackel2017semantic3d} & Terrestrial Laser Scanner (TLS) &    -   & Outdoor       &  4,000M   &    Semantic segmentation \\
		\hline
		Paris-Lille-3D\cite{roynard2018paris} & Velodyne HDL-32E (MLS)   & 1940 m &   Outdoor    &  143.1M    
		& Instance $\&$ semantic segmentation\\
		\hline
		\textbf{Campus3D (Ours)}  & \textbf{UAV photogrammtry} &   \textbf{1.58 $\times 10^6$  \scalebox{0.7}{$\textrm{m}^2$} }   &   \textbf{Outdoor }    &  \textbf{937.1M}     &\textbf{Hierarchical semantic segmentation}\\
		&             &       &             &              &    \textbf{Instance segmentation}   \\
		\bottomrule
	\end{tabular}%
	\vspace{-0.2cm}
	\label{tab:comData:collection}%
\end{table*}%

\section{Related Work} \label{sec:rw}
In this section, we firstly review the existing 3D scene-based point cloud datasets and compare our dataset with them in detail below. Based on the application area, we briefly divide the existing datasets into two  categories:  (1) indoor scene datasets and (2) outdoor scene datasets. A summary of comparison between Campus3D and the widely-used datasets is provided by Table \ref{tab:comData:collection}, and additional comparisons in terms of annotation are provided in the supplementary document. Secondly, we briefly review the existing deep neural models for point cloud segmentation.

\textbf{Indoor Dataset.} Indoor scene understanding is an active research area, and many datasets have been reported in  literature \cite{armeni2017joint,armeni20163d,chang2017matterport3d,dai2017scannet,hua2016scenenn,xiao2013sun3d,silberman2011indoor,silberman2012indoor,song2015sun}.  These datasets are usually generated by RGB-D images which can be easily got by cheap sensors (e.g., Microsoft Kinect).  
Early datasets NYUv2 \cite{silberman2012indoor}, SUN3D \cite{xiao2013sun3d} and SUN RGB-D \cite{song2015sun} were annotated by either polygons in 2D \cite{silberman2012indoor,xiao2013sun3d,song2015sun} or bounding box in 3D \cite{song2015sun}, of which the information for 3D scene reconstruction (e.g., semantic segmentation, surface reconstruction, meshes, etc.) is limited.  Recently released indoor scene datasets \cite{armeni20163d,chang2017matterport3d,hua2016scenenn,dai2017scannet,armeni2017joint}  contain  more information.  For instance,  
ScanNet \cite{dai2017scannet} supplies estimated camera parameters, surface segmentation, textured meshes and semantic segmentations; 
however, comparing with the proposed  photogrammetry dataset Campus3D,  these datasets generated by RGB-D sensors have their limitations of short measurement range and sensitivity to the sunlight's infrared spectrum \cite{hackel2017semantic3d}.  These natural limitations prevent the RGB-D datasets from applications of outdoor environment understanding.  

\textbf{Outdoor Dataset.} Several outdoor scene 3D datasets \cite{hackel2017semantic3d,roynard2018paris,serna2014paris,vallet2015terramobilita} are released in recent years.   These datasets are generated via either MLS \cite{roynard2018paris,serna2014paris,vallet2015terramobilita,behley2019semantickitti} or TLS \cite{hackel2017semantic3d}.  Points generated by the LiDAR are the raw output of the laser scanner, which are of  high quality and large scales. The MLS point cloud datasets are always annotated with rich traffic elements to push the frontier of the autonomous driving field. One notable MLS point cloud dataset is a part of KITTI  which was constructed by Geiger et al. \cite{geiger2012we, geiger2013vision} and generated from 6 hours of traffic scenarios.  Based on it,  a point cloud dataset, semanticKITTI, has been proposed  recently for outdoor semantic scene understanding \cite{behley2019semantickitti}.   However, different from our dataset collected by UAV imagery, LiDAR devices always suffer from occlusions thus lack for a holistic view of the scene.  

\textbf{Deep Segmentation Model.} Semantic segmentation and instance segmentation are the major scene understanding tasks concerned by reconstruction. As the pioneering work PointNet and PointNet++ proposed by Qi et al. \cite{qi2017pointnet, qi2017pointnet++}, point-based deep neural models come to widely-studied in point cloud segmentation field since it can directly process the point cloud. Categories of point-based deep learning models mainly include feature pooling models \cite{qi2017pointnet, qi2017pointnet++,zhao2019pointweb, hu2019randla}, convolution-based models \cite{li2018pointcnn, thomas2019kpconv, wang2018deep}, graph-based models \cite{wang2019dynamic, landrieu2018large,wang2019graph} and attention-based models \cite{xie2018attentional,yang2019modeling}.  Although most of these models are proposed for a single task, they can be involved in multitask learning. The examples are PointNet++ in ASIS \cite{wang2019associatively} and PointNet in JSIS3D \cite{pham2019jsis3d}, which jointly learn instance embedding and semantic labeling in one structure. To jointly learn semantic labeling on multiple granularity levels, we propose our modification of point-based models fitting for multitask learning. The PointNet++ is applied as backbone since its general structure and high compatibility to multitask \cite{wang2019associatively,wang2018sgpn}.

\section{Campus3D Dataset}
We note that the Campus3D is online accessible. 
Not only data can be downloaded there but also online interactive visualization and Github link for source codes are provided.  

\subsection{Data Acquisition}
The point cloud of Campus3D dataset was constructed by the technique of Structure from Motion with Multi-View Stereovision (SfM-MVS) \cite{westoby2012structure} on UAV images.  Here we briefly describe our workflow for obtaining the dataset.  Firstly, we flew drones over all areas and took images with exact GPS coordinates. The device to capture imagery was DJI Phaton 4 Pro drones equipping cameras with a 1-inch 2 MP CMOS sensors, and the drone flight planning mobile apps used in our application were DJI GS Pro and Pix4D Capture.  Then the points would be generated by photogrammetry processing and registration from captured images and coordinates using Pix4D as SfM-MVS software.

In image collection process, we applied two types of flight routing strategies for UAV photography: (1) grid and (2) circular, which were accessible in the drone flight planning mobile apps. 
For relatively high buildings, we applied multiple circular flights at different height levels. During UAV image capturing, the drone were flown when the clear view was guaranteed by the weather. More detailed settings can be found in the supplementary document.
	
\subsection{Data Annotation}
To present more complicated geometric features, we annotated the point cloud with point-wise labels.  In general, there are two approaches to perform 3D point-wise annotation: (1) label the pre-segmented clusters in 3D and (2) label the projected 2D image and assign labels to 3D points. Our strategy follows the second approach and performs a two-level of 2D projection segmentation, which avoids inherent error induced by pre-segmenting methods and lack of details in 2D projections of stationary angles. Initially, we divided the annotation tasks into hierarchical stages from coarse-grained label to fine-grained label. In each stage, annotation was firstly done by 2D polygon partitions in three orthogonal view-angles. To refine the details, the obtained 3D partitions were then pruned in user-defined rotation angles. All the tasks were completed by opensource software CloudCompare \cite{CC} and its add-on functions.  Multiple annotators were hired to perform above labeling task after taking training courses for days. To ensure the accuracy and consistency of annotations, we divided annotators to several groups, and work on labeling and verifying, respectively, for each stage. 
We require that every point is labeled at least three times by different annotators and verified to an exact label. 

According to CityGML \cite{kolbe2005citygml}, objects from urban scene are modeled in different granularity levels defined by the LoD,  
which can cope with applications in different scales. Motivated by this concept, the category labels used in the Campus3D are constructed as a hierarchical structure with various granularity levels and displayed by Figure  \ref{Fig:label_structure}. The hierarchies of the structure can work similarly to the LoDs. 
Each label is formed based on two criteria:  (1) semantic attribute and (2) geometrical attribute. They may mutually assist each other to parse the points into refined parts. For example of both ``\textit{roof}" and ``\textit{driving\_road}" with plane structure,  they are difficult to be distinguished in geometric features but need to be separated due to the semantic difference and practical function.  All labels are self-explanatory except for the following ones.   And we provide explanations for them:  (1) ``\textit{unclassified}" refers to unrecognized or over-sparse regions.  Instead of removing these data-points, this category is set for reserving the completeness of dataset;  (2) ``\textit{path\&stair}'' is only for pedestrians while ``\textit{driving\_road}'' is only for vehicles;  (3) ``\textit{artificial\_landscape}" is referring to man-made landscape such as artificial pool while ``\textit{others}" represents some individual objects because there do not exist enough instances to group them as a new category.  All the labels are defined in a rigid way for consistency of annotation.  

\subsection{Parsing and Statistics} 
To label the raw point clouds,  we propose a hierarchical parsing method for decomposing the data into individually labeled points, which is naturally generated by the hierarchical annotation in previous section.  The resulting Campus3D dataset can fulfill multiple tasks.  We firstly divide the entire dataset into six identified regions: FASS, FOE, PGP, RA, UCC and YIH according to their architecture styles and functions. A descriptive summary of points of these six regions is given by Table \ref{tab:areah}. 

\begin{table}[h]
\vspace{-0.3cm}
	\caption{Area,  mean height, points and points per area of data points of each region.}
	\footnotesize
	\centering
	\vspace{-0.2cm}
	\begin{tabular}{l|rrrr}
		\hline
		Region  & Area  & Mean  & \# of points & \# of points per \\
		&   ($\textrm{m}^2$)  &      height (m) &    &  area ($\textrm{m}^2$)\\
		\hline
		FASS  & 276,746 & 48.74  & 114,599,515 & 414.10 \\
		FOE   & 247,924 & 49.76 & 34,347,821 & 138.54 \\
		PGP   & 277,468 & 50.19 & 29,595,347 & 106.66\\
		RA    & 365,537 & 61.62 & 54,446,114 & 148.95\\
		UCC   & 127,572 & 30.09 & 333,404,689 & 2613.46\\
		YIH   & 284,775 & 42.08 &  354,491,876 & 1244.81\\
		\hline
	\end{tabular}%
	\label{tab:areah}%
	\vspace{-0.3cm}
\end{table}

\begin{figure}[htbp]
	\centering		
	\includegraphics[scale = 0.45]{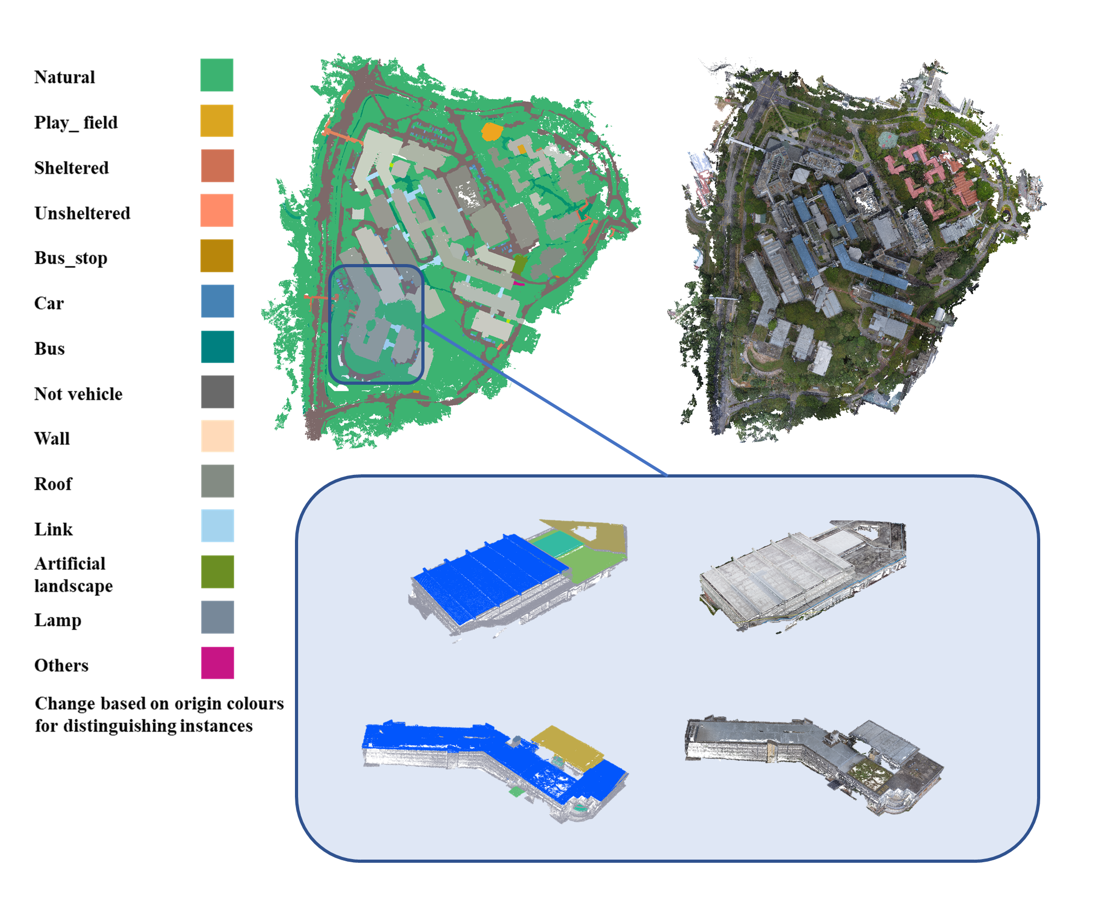} 
	\caption{Instance segmentation of building and roof. Upper left: annotated ground truth instance; upper right: raw point cloud. The bottom is a zoomed-in example of instance annotation and different colors represent different roof pieces. And there are two building instances.}
	\label{Fig:examInsSeg}
	\vspace{-0.5cm}
\end{figure}

Due to the hierarchical annotation strategy,  class labels of the Campus3D  can be defined by a tree-like structure. 
Based on the this structure,   
the coarse-grained level data can be simply obtained by merging their sub-class data including all leaf node, which is flexible for multi-level tasks. For example,  class ``\textit{building}'' data could be obtained by merging ``\textit{wall}" and ``\textit{roof}" data.   
After labeling each point by a hierarchical class tree, we performed instance labeling for each countable class, which may benefit 3D model reconstruction and scene understanding. For instance, to boost the LoD of the building model,  it is  necessary to distinguish various planar pieces from a roof.  Figure \ref{Fig:examInsSeg} illustrates this parsing.   We also note that more descriptive statistics of class and instances are provided in the supplementary document. 

\subsection{Data Preprocessing}
To practically perform the machine learning algorithms on the data, we need data simplification on point cloud with consideration of imbalanced density and processing efficiency. 
We provide a reduced dataset from the original points.  This reduced dataset is voxelly sampled from the original dataset with a sampling size of 0.15 meter.  The sampling method thins the data points and also inhibits the imbalanced distributions of points among different regions and instances, which is caused by the varies of morphology.  Moreover, the 0.15m sample size can keep the smallest object in the whole campus. We term this dataset as Campus3D-reduced.   
Note that all experimental studies, scene understanding tasks and benchmarks in this paper are run on the Campus3D-reduced.  Table \ref{tab:split} shows the training, validation and test splits. This splitting makes sure that training set and test/validation set have all types of instances.  And the performance of the class ``\textit{unclassified}'' is not included in current study, which follows the convention in this arena \cite{li2018pointcnn, qi2017pointnet, qi2017pointnet++}. 

\begin{table}[h]
	\caption{Training, validation and test set splitting}
	\centering
	\begin{tabular}{l|c|c|c}
		\hline
		&  Training   &  Validation & Test \\
		\hline
		Region  & FASS, YIH, RA, UCC & PGP & FOE  \\
		\hline
	\end{tabular}%
	\label{tab:split}%
\vspace{-0.4cm}
\end{table} 

\begin{figure}[t]
	\centering		
	\includegraphics[width= 1.0\linewidth]{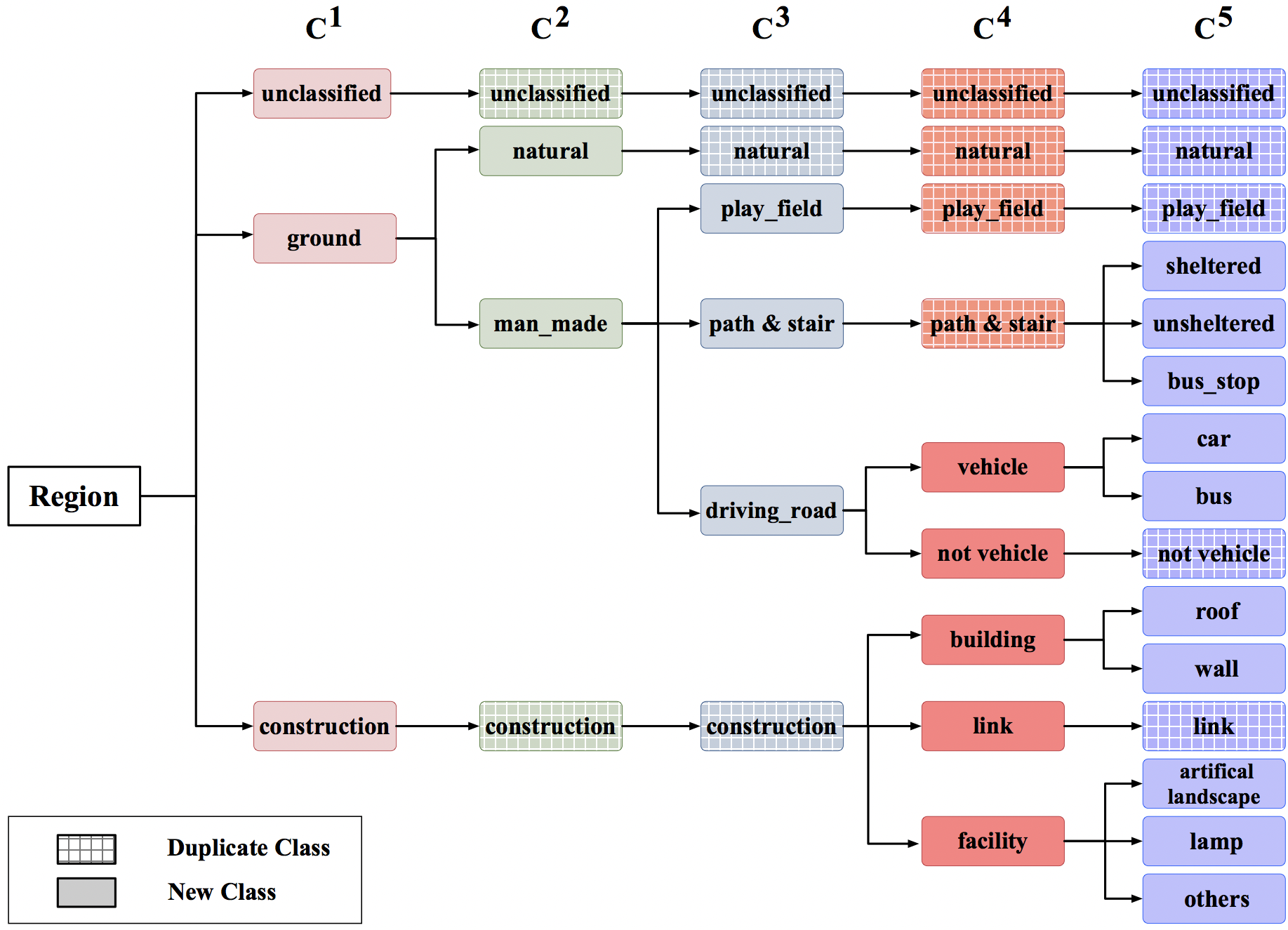} 
	\caption{Label tree $\mathcal{T}$: each internal or leaf node with solid fill represents a class; the class name of each  is inside each node.   Each data point of Campus3D dataset is annotated by a path of the tree with solid nodes. e.g.  construction -> building -> wall.  For entirely partitioning the data in each level, some nodes are duplicates of its parent nodes which are filled by grids. The tree has five ($H=5$) granularity levels:  ${C}^1 = \{\textit{unclassified}, \textit{ground}, \textit{construction}\}, \cdots,  {C}^5=\{\textit{unclassified}, \textit{natural},\cdots, \textit{others}\}$.}  
	\label{Fig:label_structure}
\vspace{-0.4cm}
\end{figure}

\section{Hierarchical Learning} \label{sec:hl}
In order to learn on hierarchical annotations of our dataset, we construct a five-level label tree displayed by Figure \ref{Fig:label_structure},  where labels in each hierarchy can completely partition the entire dataset. In that case, each point possesses five parallel semantic labels, learning of which can be consider as a multi-label segmentation tasks.
Compared with single label learning, the key problem towards hierarchical multi-label learning is how to leverage the relationship among hierarchies, while the hierarchical structure of labels should be kept. Therefore, we propose a simple yet effective framework, which includes a multi-task learning network and an  ensemble process to maintain hierarchical structure.  Before the methodology, we first proceed to the problem and performance metrics. 

\subsection{Problem and Metric Description} 
Let $(\mathbb{C}, \leq_{\eta})$ represent the class hierarchy, where $\mathbb{C}$ is a set of classes and $\leq_{\eta}$ is a partial order representing the superclass relationship. For any $c_1, c_2 \in \mathbb{C}$, $c_2 \leq_{\eta} c_1$ if and only if $c_1$ is a superclass of $c_2$ or $c_1 = c_2$.
Data point $i$ with hierarchical annotation is denoted as  $(X^i, S^i)$ with $X^i \in \mathcal{X}=\mathbb{R}^D$ and $S^i$ is a maximal chain of $\mathbb{C}$.  The problem of such label is that length of the label set $|S_i|$ is not coherent from point to point. To construct a multi-label with coherent length,  we further extend the definition of hierarchical learning by allowing duplication.  We first notate the set of all  maximal elements in $\mathbb{C}$ by $C_{max}$ and the set of all minimal elements $C$ by $C_{min}$. Note that $C_{max}$ and $C_{min}$ both belong to $\mathcal{A}_\mathbb{C}$, the set of all antichains in  $\mathbb{C}$. We define a relationship to compare the any two antichains named parent antichain:

\begin{definition}[Parent Antichain]
For two distinct sets $C_c$, $C_p \in \mathcal{A}_\mathbb{C}$, if $\forall c_j \in C_c$, $\exists c_j' \in C_p$ let $c_j\leq_{\eta}c_j'$,  then $C_p$ is called a parent antichain of $C_c$ with notation $C_c \prec_{\eta} C_p$.
\end{definition}
Then we can obtain a sequence of antichains (sets) between $C_{\textrm{min}}$ and $C_{\textrm{max}}$ if  $C_{\textrm{min}} \neq C_{\textrm{max}}$, namely, $\{C^h\}_{h=1}^H=(C^1= C_{\textrm{max}}, C^2,$ $\ldots, C^H = C_{\textrm{min}})$ with length $H=\max |S_i|$ that $C^H \prec_{\eta} C^{H-1}, \ldots, $ $\prec_{\eta} C^1$ and $\cup_{h=1}^H C^h = \mathbb{C}$. 
Based on the sequence, a tree $\mathcal{T}$ can be constructed and displayed in Figure \ref{Fig:label_structure}. The nodes in $h^{th}$ layer of the tree can be associates with $C^{h}$, while the edge is defined as the partial order relationship between classes. 
Now we  define the hierarchical learning problem. A dataset $\mathcal{D} = \{(X^i, Y^i)|i\in \mathbb{Z}, 1\leq i \leq N\}$, where $N$ is the number of points, $X^i \in \mathcal{X}$ and $Y^i \in \mathcal{Y} = C^1\times C^2 \times \ldots  \times C^H $.  The hierarchical learning problem is to learn a function $f$: $\mathcal{X} \mapsto \mathcal{Y}$ from a hierarchically annotated dataset $\mathcal{D}$.

Given a HL method $f(\cdot)$, the performance can be evaluated by the conventional classification measurements such as accuracy, precision, recall, etc.  However, they fail to take consistency into account which is critical for the HL.  
It is possible that a HL algorithm performs good in terms of conventional measurements, but generates highly inconsistent results violating the hierarchical relationship which are meaningless.  Therefore, we propose a new measurement and quantitatively evaluate such consistency.   Considering a solution (prediction) $Y \in \mathcal{Y}$, we  first  define the \textit{fully consistent (FC)} for a solution at Definition 4.1.    The set of all FC solutions is denoted as $\mathcal{Y}^{\textrm{FC}}$ ($\mathcal{Y}^{\textrm{FC}} \subset \mathcal{Y}$), and it includes all paths from root to leaf nodes in tree $\mathcal{T}$.   Based on it,  we propose, \textit{consistency proportion (CP)},  to measure the consistency degree for solution $Y^i$.   The CP value is between 0 and 1, and being one represents a  FC solution.    Then, for a set of solutions $\{Y^1, \ldots, Y^N\}$,   the \textit{consistency rate (CR)} is defined with parameter $\alpha$ being the desired consistency level for each solution.   

\begin{definition}[Fully Consistent]
\label{Methodology:def:fc}
Solution $Y = (y_1, \ldots, y_H)$ $\in \mathcal{Y}$ is defined as fully consistent (FC) if it satisfies $y_H \leq_{\eta} y_{H-1} \leq_{\eta} \ldots \leq_{\eta} y_1$.
\end{definition}

\begin{definition}[Consistency Proportion]
	The consistency proportion (CP) of $Y^i = (y^i_1, \ldots, y^i_H)$  is defined as:
	\begin{equation}
		\textrm{CP} (Y^i)  = \frac{\max\limits_{(y_1,\ldots, y_H) \in \mathcal{Y}^{\textrm{FC}}} \sum^H_{h=1} \mathbb{1} (y^i_h = y_h)} {H},
	\end{equation}
	here $\mathbb{1}(x) = 1$ if $x$ is True; 0 otherwise. 
\end{definition}

\begin{definition}[Consistency Rate]
	The consistency rate (CR) with CP level $\alpha$ for $\{Y^1, \ldots, Y^N\}$ is:
	\begin{equation}
		\textrm{CR}_{\alpha} = \frac{1}{N} \sum^N_{i=1} \mathbb{1} [\textrm{CP}(Y^i)  \geq \alpha].
	\end{equation}
	Here $\alpha$ is a threshold parameter and $0 \leq \alpha \leq 1$. 
\end{definition}

\begin{figure*}[t]
	\centering		
	\includegraphics[scale = 0.15]{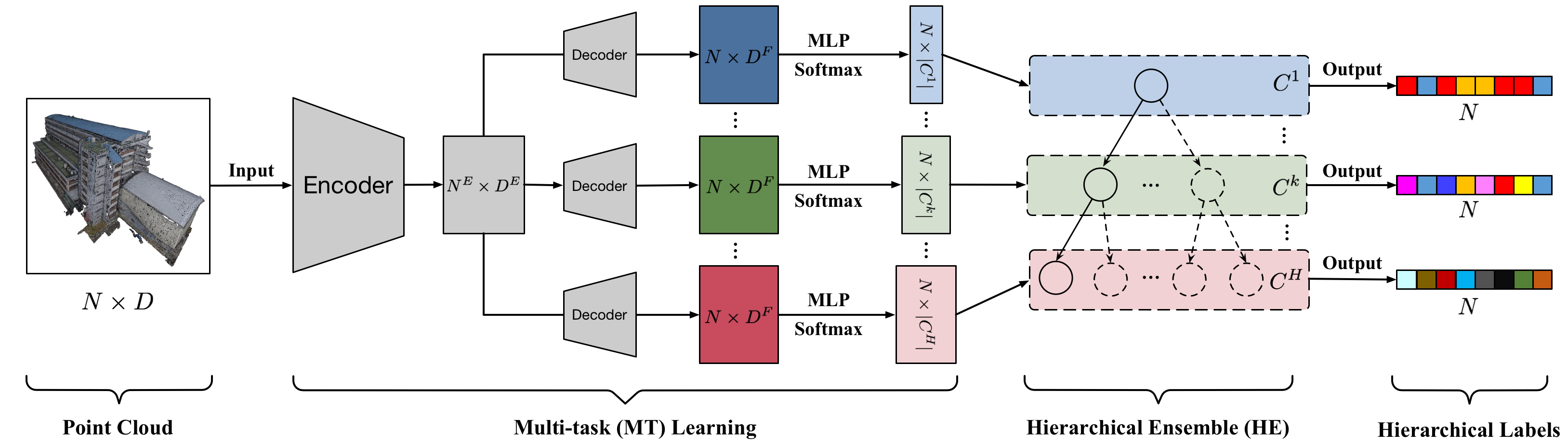} 
	\caption{The framework of our method is divided into two stages: Multi-task Learning (MT) and Hierarchical Ensemble (HE). Point cloud data is fed as input. After feature extraction through the shared encoder, the features are decoded into multiple heads, and the predicted distributions at different granularity levels are obtained via MLP layers. Then, the HE stage utilizes hierarchical relationship to gain the final hierarchical labels. The width of the model depends on the hierarchical levels, where the $k^{th}$-level represents the middle parts. $N^E \times D^E$ is the size of embedding from the encoder. The MLP output dimension $|C^k|$ depends on the number of labels on the $k^{th}$ granularity level.}
	\label{Fig:proposedmethod}
\end{figure*}

\subsection{Methodology}
We propose a two-stage framework to the HL (see Figure \ref{Fig:proposedmethod}): (1) multi-task learning (MT) and (2) hierarchical ensemble (HE).  

\textbf{Multi-task Learning (MT).} The main structure of MT networks contains a shared encoder and multiple parallel decoders with classification heads. To practically perform the MT, we utilized the feed forward architecture of PointNet++ \cite{qi2017pointnet++}. Specifically, an feature map $N \times D$ of point cloud with size $N$ and feature dimension $D$ is fed as input. Then the shared encoder encodes them as embedding.  Such embedding is then decoded parallelly into $N \times D^{F}$ by $H$ decoders for $H$ granularity levels.  Decoder $h$ computes the likelihood distributions of classes ($C^h$) at granularity level $h$  for each data point.  
The loss of MT method is the sum of the losses of its branches,
\begin{equation}
\mathcal{L}_{\textrm{MT}} =\mathcal{L}_{\textrm{prediction}} +\mathcal{L}_{\textrm{consistency}}
\end{equation}
where the prediction loss $\mathcal{L}_{\textrm{prediction}}$ is the weighted average of the cross entropy losses of $H$ levels.  And it is formulated as,
\begin{equation}
\mathcal{L}_{\textrm{prediction}} =\sum ^{H}_{h=1} \beta_{h} \cdot L^{h}_{\textrm{prediction}}.
\end{equation}
Here, for granularity level $h$,  $L^{h}_{\textrm{prediction}}$ and $\beta_h$ are  the  cross-entropy loss and weight respectively. 
The consistency loss is served as a regularization term to maintain the consistence structure of predicted distributions
\begin{equation}
\mathcal{L}_{\textrm{consistency}} = \sum ^{H-1}_{h=1} \gamma _{h}  \sum_{(y_h, y_{h+1}) \in \textrm{PC}^h}  \left[ P^{h+1} \left( y_{h+1} \right) - P^{h} \left( y_h \right) \right]^{2}_{+} \ 
\end{equation}
here $\textrm{PC}^h = \{(y_h, y_{h+1}) | (y_1,\ldots, y_H) \in \mathcal{Y}^{\textrm{FC}}\}$, 
where $\gamma^{h}$ is the loss weight of $h^{th}$ level and $P^{h}(\cdot)$ is the predicted likelihood distribution over class set (antichain) $C^h$.  By Definition 4.2, given FC solution $(y_1,\ldots, y_H)$,   $y_h$ is the superclass of or same as $y_{h+1}$.  
This loss is the sum of the losses of all the parent-child pair in tree $\mathcal{T}$, which is to keep a smaller prediction score $ P{^{h+1}}\left( y_{h+1} \right)$ than its parent score $ P{^{h}}\left(  y_{h} \right)$ such that consistency is reserved.
To investigate the effectiveness of the consistency loss, a loss without consistency loss branch named \textbf{MT}$_{\textrm{nc}}$ is tested to perform hierarchical semantic segmentation as ablation study,
\begin{equation}
\mathcal{L}_{\textrm{MT}_{\textrm{nc}}} =\mathcal{L}_{\textrm{prediction}}.
\end{equation}

\textbf{Hierarchical Ensemble (HE).}  The HE is a post-processing method for initial predicted results.   It computes the weighted sum of likelihood scores over all the root-to-leaf paths in tree $\mathcal{T}$.  The path associated with largest score is the final predicted solution. It is formulated as equation (\ref{eq:HE}). 
Note that solutions generated by HE are FC and the CR (and/or CP) value is 1.  
\begin{equation} \label{eq:HE}
Y_{\textrm{HE}} =\arg\max_{ (y_1,\ldots, y_H) \in \mathcal{Y}^{\textrm{FC}}} \sum ^{H}_{h=1} P^{h}\left( y_h \right)
\end{equation}

In order to perform comparison analysis, we also apply  a multi-classifier (\textbf{MC}) method which does not leverage the mutual relationship across levels, and  only trains an independent segmentation classifier for each granularity level.    And $H$ classifiers are trained and evaluated separately. It 
performs conventional segmentation $H$ times for the dataset based on PointNet++.  
Futhermore, a variant of the proposed two-stage method, MC+HE is also investigated, which uses the HE to post-process outputs of the MC.

\subsection{Experimental Results} \label{sec:tb:hss}
Based on the class label tree given by Figure \ref{Fig:label_structure},  we build five granularity levels ($H=5$). They are given in the first and second columns of Table \ref{tab:he:test}.  PointNet++ \cite{qi2017pointnet++} is used as backbone. We set $\beta_{i}$=1 and $\gamma_{i}=0.05$ with $i\in \{1,\cdots, H\}$, and more detailed settings are present in the supplementary material. We apply $\textrm{CR}_1$ ($\alpha=1$), intersection-over-union (IoU) as well as overall accuracy (OA) for performance analysis.  
\begin{table}[htbp]
\caption{Test results (OA\%) for different HL methods.}
\begin{tabular}{llllll}
\hline
	\multirow{2}{*}{Method}  & \multicolumn{5}{c}{Granularity Level} \\ 
           & $C^1$    & $C^2$    & $C^3$    & $C^4$    & $C^5$  \\ 
    \hline
    MC    & 85.3  & 79.5  & 78.3  & 76.3  & 74.0  \\
    MC+HE & 89.1  & 81.4  & 79.9  & 77.9  & 73.5  \\
    MT$_{\textrm{nc}}$ & 90.2  & 82.2  & 80.9  & 78.8  & 74.6  \\
    MT    & \textbf{90.7 } & \textbf{83.1 } & \textbf{81.7 } & 79.8  & 75.2  \\
    MT+HE & \textbf{90.7}  & \textbf{83.1 } & \textbf{81.7 } & \textbf{80.0 } & \textbf{75.4 } \\
    \bottomrule
    \end{tabular}%
    \vspace{-0.2cm}
  \label{tab:oa:test}
\end{table}%

\textbf{Comparisons between different HL methods.} After removing points with ground-truth label of ``\textit{unclassified}'' (unlabeled),  for each class,  the intersection and union  sets of predicted point set and ground-truth are generated; then the IoUs are computed as the ratio of intersection set cardinality to that of the union. And the OA is computed as proportion of correct predictions to total points. Results for test set results  are presented in  Table \ref{tab:oa:test} 
and Table \ref{tab:he:test}.  There are several observations:  (1) in terms of average IoU and OA of the granularity level, it decreases significantly with granularity level changing from  $C^1$ to $C^5$.  It indicates that the difficulty of the problem  increases as the label instances become small and distributed sparsely;  (2) the performance of MC + HE is better than that of MC only for most cases;  (3) overall, the HL methods  (i.e., MT + HE, MT, MT$_{\textrm{nc}}$, MC + HE) taking hierarchical labels into account perform better than the MC without considering them.  These observations demonstrate that hierarchical labels help and enhance the performance.  

\begin{figure}[htbp]
	\centering		
	\includegraphics[scale = 0.55]{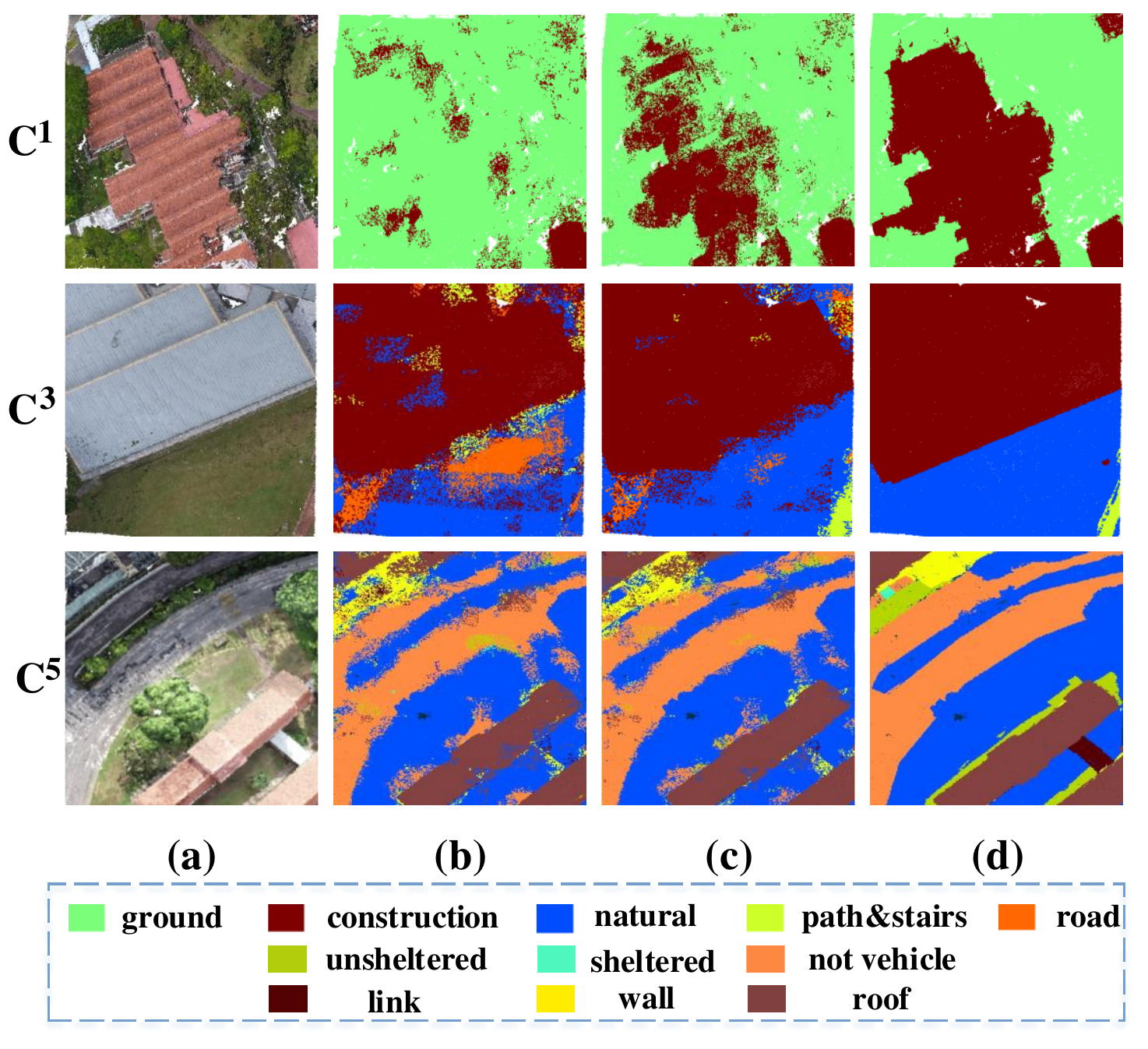} 
	\caption{Visualization of hierarchical segmentation results. (a): raw point cloud;   (b) and (c) are MC and MT  results, respectively, in $C^1$, $C^3$ and $C^5$ levels; 	(d) : ground truth label.} 
	\label{Fig:geo_amb}
\vspace{-0.5cm}
\end{figure}

One possible reason of the better performance by the HL method is that the inherent relation among label layers provide additional geometrical information for semantic segmentation.  
A visual illustration is given by Figure \ref{Fig:geo_amb}.   The MC semantic segmentation on the level $C^5$ and $C^3$  without other level information results in that ``\textit{roof}'' is wrongly recognized as driving road (i.e.``\textit{road}''or ``\textit{not vehicle}'') or natural ground ``\textit{natural}''  (see $C^5$ (b) and $C^3$ (b)  of Figure \ref{Fig:geo_amb}).  We found that they are geometrically similar but semantically different.  Here we first define this phenomenon as  \textit{geometric ambiguity}:  points with similar geometric features but significantly different semantic labels are wrongly classified to the same semantic class. As indicated by the result of MT ((c) column of Figure \ref{Fig:geo_amb}), hierarchical and multiple annotation can ameliorate  this phenomenon.   For the instance of $C^1$ level in Figure \ref{Fig:geo_amb}, points on roof belonging to ``\textit{construction}'' are easily recognized as ``\textit{ground}'' by the MC, while the MT framework is able to segment them correctly by leveraging information from finer levels.

\begin{table}[htbp]
\vspace{-0.2cm}
  \caption{Test results (class IoU\%) for HL methods}
\centering
\resizebox{0.95\columnwidth}{!}{
\begin{tabular}{clccccc}
\toprule
 Granularity &  & \multicolumn{5}{c}{Method} \\ 
 \cline{3-7} 
Level & \multirow{-2}{*}{Class} & MC & MC+HE & $\textrm{MT}_{\textrm{nc}}$ & MT & MT+HE \\
\midrule
    \multirow{2}{*}{$C^1$} & ground & 78.9  & 83.5  & 84.9  & \textbf{85.5 } & \textbf{85.5 } \\
                           & construction & 67.4  & 75.5  & 78.2  & \textbf{79.1 } & 79.0  \\
    \hline
    
    \multirow{3}{*}{$C^2$} & natural & 66.8  & 69.4  & 71.5  & \textbf{71.9} & \textbf{71.9} \\
                           & man\_made & 52.6  & 54.7  & 53.7  & \textbf{54.8} & 54.7  \\
                           & construction & 72.9  & 75.5  & 77.0  & \textbf{79.1} & 79.0  \\
    \hline
    
    \multirow{5}{*}{$C^3$} & natural & 67.1  & 69.4  & 71.8  & \textbf{71.9 } & \textbf{71.9 } \\
                           & play\_field & 0.3   & 0.3   & 0.7   & \textbf{3.0 } & 2.2  \\
                           & path\&stair & 7.2   & 8.2   & \textbf{7.8 } & 0.0   & 0.0  \\
                           & driving\_road & 49.9  & 52.0  & 51.2  & \textbf{52.5 } & 52.3  \\
                           & construction & 74.1  & 75.5  & 77.7  & \textbf{79.1 } & 79.0  \\
    \hline
    
    \multirow{8}{*}{$C^4$}  & natural & 67.8  & 69.4  & \textbf{71.9 } & 71.8  & \textbf{71.9}  \\
                            & play\_field & 0.8   & 0.3   & 0.5   & 1.9   & \textbf{2.2 } \\
                            & path\&stair & 7.8   & \textbf{8.2}   & 7.9  & 0.0   & 0.0  \\
                            & vehicle & 34.5  & \textbf{38.7 } & 36.7  & 36.4  & 36.6  \\
                            & not vehicle & 48.6  & 51.1  & 50.1  & 51.1  & \textbf{51.3 } \\
                            & building & 70.1  & 72.1  & 74.0  & 75.7  & \textbf{76.0 } \\
                            & link  & 2.1   & 2.2   & \textbf{3.7 } & 0.5   & 0.5  \\
                            & facility & 0.0   & 0.0   & \textbf{0.1 } & 0.0   & 0.0  \\
    \hline
    
    \multirow{14}{*}{$C^5$}  & natural & 71.0  & 69.4  & \textbf{72.2 } & 71.7  & 71.9  \\
                             & play\_field & 1.6   & 0.3   & 2.0   & 1.8   & \textbf{2.2 } \\
                             & sheltered & \textbf{10.7 } & 10.4  & \textbf{10.7}  & 1.4   & 0.0  \\
                             & unsheltered & 4.4   & 4.4   & \textbf{4.7 } & 0.1   & 0.0  \\
                             & bus\_stop & 0.4  & \textbf{0.5}   & 0.1   & 0.0   & 0.0  \\
                             & car   & \textbf{40.7 } & 39.9  & 38.5  & 37.0  & 36.6  \\
                             & bus   & 0.0   & 0.0   & 0.0   & 0.0   & 0.0  \\
                             & not vehicle & 50.6  & 51.1  & 50.1  & 51.0  & \textbf{51.3 } \\
                             & wall  & 57.3  & 56.0  & 57.2  & 57.1  & \textbf{57.5 } \\
                             & roof  & 58.6  & 57.2  & 60.4  & 61.1  & \textbf{61.3 } \\
                             & link  & 3.5  & 2.2   & \textbf{3.6}   & 0.5   & 0.5  \\
                             & artificial\_landscape & 0.0   & 0.0   & 0.0   & 0.0   & 0.0  \\
                             & lamp  & 0.0   & \textbf{0.1 } & 0.0   & 0.0   & 0.0  \\
                             & others & \textbf{0.2 } & 0.0   & \textbf{0.2 } & 0.0   & 0.0  \\
    \bottomrule
    \end{tabular}%
}
  \label{tab:he:test}%
  \vspace{-0.2cm}
\end{table}%

\begin{figure}[htbp]
\label{fig:CR}
    \centering
	\includegraphics[scale = 0.35]{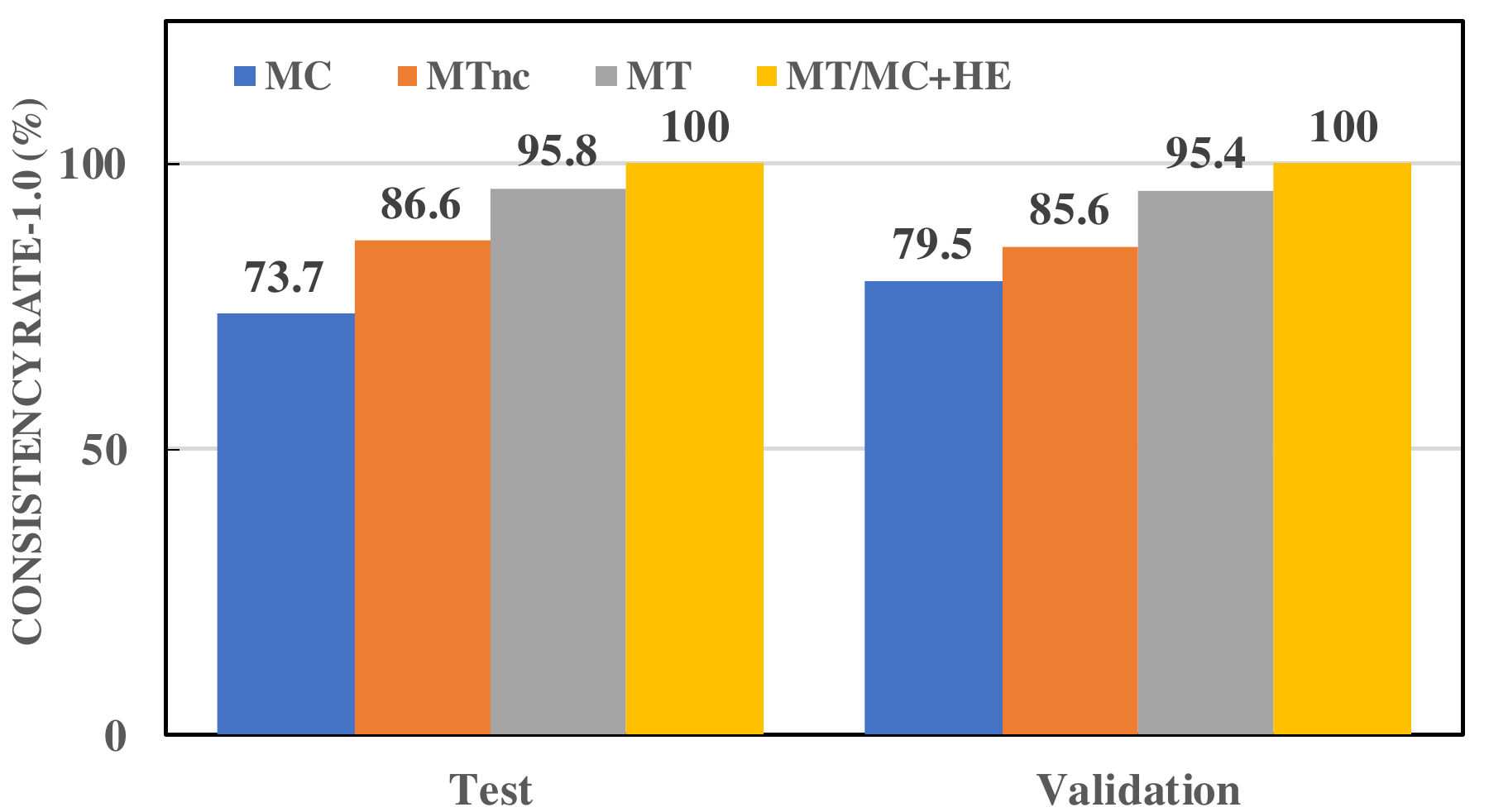} 
	\caption{Results of $\textrm{CR}_{1}$ (\%) using different HL methods.}
	\label{Fig:consistency_bar}
\vspace{-0.35cm}
\end{figure}

\textbf{Insights of Consistency Rate.} As a metric evaluating consistency of hierarchical relationship, the result of CR$_1$ reveals that our framework leverages the mutual assistance among hierarchies. From Figure \ref{Fig:consistency_bar}, the $\textrm{CR}_1$ of MT is over around 15\% than that of MC which ignores the hierarchical annotation.  It suggests that MT learning may correct the results in certain level according to features from other granularity levels. On another hand, with comparison of MC and MC+HE results from Table \ref{tab:he:test} and Table \ref{tab:oa:test}, HE also boosts the performance by maintaining hierarchical relationship forcibly, while this boosting is not significant from MT to MT+HE. The results of $\textrm{CR}_{1}$ quantitatively explain why HL methods can effectively address the geometric ambiguity discussed above.  

\textbf{Effectiveness of Consistency Loss.}  Performance differences between MT and MT$_\textrm{nc}$ in Figure \ref{Fig:consistency_bar} and  Table \ref{tab:oa:test} demonstrate the effectiveness of the proposed consistency loss. With it, MT can significantly restrain the hierarchical violations in segmentation results,  while MT$_{\textrm{nc}}$, ignoring it,  results in around 10\% decrease in terms of $\textrm{CR}_1$.  Moreover, MT also performs better than MT$_{\textrm{nc}}$  in terms of OA (see Table \ref{tab:oa:test}).  

\section{Hierarchical Scene Understanding Tasks and Benchmarks}
In this section, we apply the HL framework for scene understanding and build benchmarks on two tasks of hierarchical semantic segmentation and instance segmentation.  We first investigate the effectiveness of sampling methods for feature learning on large-scale point cloud datasets.  

\subsection{Sampling Methods} \label{sec:sm}
For scene based point cloud datasets, sampling is necessary for feature learning because of requirements of efficiency and fixed size of model input.  Typical sampling strategies are uniform sampling  and farthest point sampling (FPS) \cite{qi2017pointnet}. Here  we do not apply farthest point sampling (FPS) due to its computational inefficiency.  The simplest sampling method is to randomly pick a fixed size of points with  uniform distribution.  However, randomly sampling a small size of points from large point cloud data would induce considerable randomness into the samples, which may lead to failure of training process.   Therefore, to conduct less bias and learnable sampling,  we experimented two variations of uniform sampling, the details of which are presented in the supplementary document: 
(1) $l$-$w$ random block sampling ($l$-$w$ RBS), and (2) random centered K nearest neighor (RC-KNN).  Given a set of points $\mathcal{D}_{x} = \{(x^1_{1}, x^1_{2}, x^1_{3}), (x^2_{1}, x^2_{2}, x^2_{3}), \ldots, (x^n_{1}, x^n_{2}, x^n_{3})\} \in \mathbb{R}^3$ ($D=3$ and three coordinates: latitude, longitude and height),   we define them and illustrate how to select $N$ ($N < n$) points from $\mathcal{D}_{x}$ as follows.

\textbf{$l$-$w$ RBS}  randomly chooses a point $P_c$ $(x^c_{1}, x^c_{2}, x^c_{3}) $ from $\mathcal{D}_{x}$ with an uniform distribution and then uniformly samples $\mathcal{D}^{\prime}_{x} = $ $ \{(x^c_{1}, x^c_{2}, x^c_{3}),$ $ (x^{i_1}_1, x^{i_1}_2, x^{i_1}_ 3), $ $\ldots, (x^{i_{(N-1)}}_1, x^{i_{(N-1)}}_2, x^{i_{(N-1)}}_3) \}$ in a $l$-$w$ block centered at $P_c$, namely, for any $1 \leq j \leq N-1$,   $x^c_{1} - \frac{l}{2} \leq x^{i_j}_1 \leq x^c_{1} + \frac{l}{2}$ and  $x^c_{2} - \frac{w}{2} \leq x^{i_j}_2 \leq x^c_2 + \frac{w}{2}$.  

\textbf{RC-KNN} randomly chooses a point $P_c$ $(x^c_{1}, x^c_{2}, x^c_{3}) $ from $\mathcal{D}_{x}$ with an uniform distribution, and then $K$ ($K = N$) nearest neighors to point $P_c$ in terms of Euclidean distance are chosen as the sampled points.    

In order to investigate the effectiveness of the above two sampling methods on the HL methods, we apply them with PointNet++ \cite{qi2017pointnet++} as the feature leaning network on our dataset.  Specifically, in each training iteration, we use either RBS or RC-KNN to select $N$ points from a randomly picked region in training set as a sample in a batch.  The sample size $N$ is set as 2048,  and block size of both $l$ and $w$  in RBS are set as 12m.   
We compute the mean IoU (mIoU) across all classes for each granularity level.  The test results 
are given in Table 
\ref{tab:sampling:test}. 
From the results, we can see that the RBS method dominates the RC-KNN method by our setting.  Note that the setting of RBS sampling is also utilized in Section \ref{sec:hl}.
\begin{table}[htbp]
\vspace{-0.2cm}
	\caption{Semantic segmentation results (mIoU\%) for RBS and RC-KNN sampling methods. }  
	\centering
	\small
	\vspace{-0.2cm}
	\begin{tabular}{lllllll}
		\toprule
		\multirow{2}{*}{Model} & \multicolumn{1}{l}{Sampling} & \multicolumn{5}{c}{Granularity level} \\
		& \multicolumn{1}{l}{Method} & \multicolumn{1}{l}{$C^1$} & \multicolumn{1}{l}{$C^2$} & \multicolumn{1}{l}{$C^3$} & \multicolumn{1}{l}{$C^4$} & \multicolumn{1}{l}{$C^5$} \\
		\midrule
		\multirow{2}{*}{MT} & RBS   & \textbf{82.3 } & \textbf{68.6} & \textbf{41.3 } & \textbf{29.7 } & \textbf{20.1 } \\
		& RC-KNN & 77.4  & 61.7  & 37.2  & 25.5  & 9.9  \\
		\midrule
		\multirow{2}{*}{MT + HE} & RBS   & \textbf{82.2 } & \textbf{68.5 } & \textbf{41.1 } & \textbf{29.8 } & \textbf{20.1 } \\
		& RC-KNN & 77.0  & 62.0  & 37.1  & 25.6  & 9.8  \\
		\bottomrule
	\end{tabular}%
	\label{tab:sampling:test}%
	\vspace{-0.4cm}
\end{table}%

\subsection{Semantic Segmentation} \label{sec:tb:ss}
In this section, we benchmark the performance on semantic segmentation task.  Three established models are applied: PointNet++ \cite{qi2017pointnet++},  PointCNN \cite{li2018pointcnn} and DGCNN \cite{wang2019dynamic}.  The sampling method used here is the RBS sampling with block size of 12m ($l=w=\textrm{12m}$).  We take the hierarchical annotation into account and apply our the proposed MT+HE method.   The mIoU across all classes for each granularity is used as the performance metric.  The results of both test and validation are given by 
Table \ref{tab:res:ss:mtche}.  

\begin{table}[htbp]
	\caption{Semantic segmentation results (mIoU\%)  for three feature learning models with HL methods. } 
	\small
	\centering
	\vspace{-0.4cm}
	\begin{tabular}{lllllll}
		\toprule
		\multirow{2}{*}{Dataset} & \multicolumn{1}{l}{Learning} & \multicolumn{5}{c}{Granularity level} \\
		& \multicolumn{1}{l}{Model} &  {$C^1$}& \multicolumn{1}{l}{$C^2$} & \multicolumn{1}{l}{$C^3$} & \multicolumn{1}{l}{$C^4$} & \multicolumn{1}{l}{$C^5$} \\
		\midrule
		\multirow{3}{*}{Validation} & PointNet++ & 79.7  & 67.0  & 43.4 & 33.4  & 21.9  \\
		& PointCNN & 86.9  & 77.4  & 51.1  & 38.0  & 27.2  \\
		& DGCNN & \textbf{88.1 } & \textbf{79.8 } & \textbf{53.0 } & \textbf{39.5 } & \textbf{29.5 } \\
		\midrule
		\multirow{3}{*}{Test} & PointNet++  & 79.7  & 67.0  & 43.4 & 33.4  & 21.9  \\
		& PointCNN & 88.6  & 78.2  & 58.4  & 41.1  & 27.2  \\
		& DGCNN & \textbf{90.9 } & \textbf{81.3 } & \textbf{61.5 } & \textbf{43.6 } & \textbf{29.1 } \\
		\bottomrule
	\end{tabular}%
	\label{tab:res:ss:mtche}%
	\vspace{-0.5cm}
\end{table}%

\subsection{Instance Segmentation} \label{sec:tb:is}
In this section, we build the instance segmentation benchmark of current dataset.   
The training, validation and test splitting still follows Table \ref{tab:split}.   We perform this task for the granularity level four ($C^4$) only, where there exists the largest number of available classes and instances among all granularity levels for training, validation and test.   The ASIS \cite{wang2019associatively} and SGPN \cite{wang2018sgpn} method were used here for the baseline evaluation.   For each class, the weight coverage (WCov) as introduced by Wang et al. in  \cite{wang2019associatively} is computed as the performance measurement.   Results for both validation and test sets are shown in Table \ref{tab:ins_res}, which shows that ASIS \cite{wang2019associatively} performs  better than SGPN \cite{wang2018sgpn}.   Note that classes ``\textit{natural}'', ``\textit{path\&stair}'', ``\textit{not vehicle}'' and ``\textit{facility}'' are not countable, thus no instance segmentation results are for them.    

\begin{table}[htbp]
	\centering
	\small
	\caption{Instance segmentation results (WCov)  for selected classes at granularity level four ($C^4$).} 
	\vspace{-0.2cm}
	\begin{tabular}{l|ll|ll}
		\toprule
		Instance class & \multicolumn{2}{c|}{Validation} & \multicolumn{2}{c}{Test} \\
		\cline{2-5}
		name & ASIS  & SGPN  & ASIS  & SGPN \\
		\midrule
		play\_field & 0.0   & 0.0   & \textbf{3.4}   & 0.0  \\
		vehicle & \textbf{34.0}  & 17.3  & \textbf{44.1}  & 32.7  \\
		building & \textbf{53.4}  & 35.7  & \textbf{40.3}  & 32.1  \\
		link  &  \textbf{13.3}  & 10.1  & \textbf{11.2}  & 8.5  \\
		\midrule
		Average &\textbf{26.4}  & 15.8  & \textbf{24.8}  & 18.3  \\
		\bottomrule
	\end{tabular}%
	\label{tab:ins_res}
	\vspace{-0.5cm}
\end{table}%

\section{Conclusion}
A well-annotated point cloud dataset with two benchmarks, Campus3D, is proposed in this paper.  It is annotated with multiple and hierarchical label for the better scene understanding and potential usage in reconstruction. We define the HL problem and propose a new measure to evaluate the consistency across granularity levels. A two-stage method MT+HE is presented to the HL. 
Experimental results demonstrate its effectiveness comparing with MC without taking multiple and hierarchical information into account.   Moreover, we investigate two sampling methods for point cloud learning with HL methods and identify RBS as the useful one.  Future users will benefit from these initial and basic explorations.  In the end, we apply established models and benchmark performance for semantic and instance segmentation for future comparisons.  
Other potential tasks can be built based on the Campus3D such as hierarchical instance segmentation and 3D model reconstruction. 

\section*{Acknowledgments}
This research is supported by National University of Singapore (NUS) Institute of Operations Research and Analytics (IORA) grant R-726-000-002-646 and National Research Foundation of Singapore grant NRF-RSS2016-004.  The authors gratefully acknowledge the data collection support of Virtual NUS team. 

\bibliographystyle{ACM-Reference-Format}
\bibliography{Campus3D}
\end{document}